\documentclass[letterpaper]{article} % DO NOT CHANGE THIS
\usepackage{aaai24}  % DO NOT CHANGE THIS
\usepackage{times}  % DO NOT CHANGE THIS
\usepackage{helvet}  % DO NOT CHANGE THIS
\usepackage{courier}  % DO NOT CHANGE THIS
\usepackage[hyphens]{url}  % DO NOT CHANGE THIS
\usepackage{graphicx} % DO NOT CHANGE THIS
\usepackage{natbib}  % DO NOT CHANGE THIS AND DO NOT ADD ANY OPTIONS TO IT
\usepackage{caption} % DO NOT CHANGE THIS AND DO NOT ADD ANY OPTIONS TO IT
\usepackage{latexsym}
\usepackage{amssymb}
\usepackage{amsmath}
\usepackage{amsthm}
\usepackage{booktabs}
\usepackage{enumitem}
\usepackage{graphicx}
\usepackage{caption}
\usepackage[utf8]{inputenc}
\usepackage{booktabs}
\usepackage[switch]{lineno}

\usepackage{colortbl}
\definecolor{myGray}{gray}{0.85}
\usepackage[dvipsnames]{xcolor}
\usepackage{subcaption}
\usepackage{subfig}
\usepackage{array,multirow}
\usepackage{pifont}

\usepackage[ruled,vlined]{algorithm2e}
\definecolor{commentcolor}{RGB}{110,154,155}   % define comment color
\newcommand{\PyComment}[1]{\ttfamily\textcolor{commentcolor}{\# #1}}  % add a "#" before the input text "#1"
\newcommand{\PyCode}[1]{\ttfamily\textcolor{black}{#1}} % \ttfamily is the code font

\newcommand{\BibTeX}{B\kern-.05em{\sc i\kern-.025em b}\kern-.08em\TeX}

\title{Representation Magnitude has a Liability to Privacy Vulnerability}
\author{
    Xingli Fang\textsuperscript{\rm 1}, 
    Jung-Eun Kim\textsuperscript{\rm 1}\thanks{Correspondence.}
}
\affiliations{
    \textsuperscript{\rm 1}Computer Science, North Carolina State University\\
}

\begin{document}

\maketitle

% \begin{abstract}
% The privacy-preserving approaches to machine learning (ML) models have made substantial progress in recent years.
% However, it is still opaque in which circumstances and conditions the model becomes privacy-vulnerable, leading to a challenge for ML models to maintain both performance and privacy.
% In this paper, we first explore the disparity between member and non-member data in the representation of models under common training frameworks. We identify how the representation magnitude disparity correlates with privacy vulnerability and address how this correlation impacts privacy vulnerability. Based on the observations, we propose Saturn Ring Classifier Module (SRCM), a plug-in model-level solution to mitigate membership privacy leakage. Via a confined yet effective representation space, our approach ameliorates models' privacy vulnerability while maintaining generalizability.
% \end{abstract}

\begin{abstract}
The privacy-preserving approaches to machine learning (ML) models have made substantial progress in recent years.
However, it is still opaque in which circumstances and conditions the model becomes privacy-vulnerable, leading to a challenge for ML models to maintain both performance and privacy.
In this paper, we first explore the disparity between member and non-member data in the representation of models under common training frameworks. We identify how the representation magnitude disparity correlates with privacy vulnerability and address how this correlation impacts privacy vulnerability. Based on the observations, we propose Saturn Ring Classifier Module (SRCM), a plug-in model-level solution to mitigate membership privacy leakage. Through a confined yet effective representation space, our approach ameliorates models' privacy vulnerability while maintaining generalizability. The code of this work can be found here: \url{https://github.com/JEKimLab/AIES2024_SRCM}
\end{abstract}

\section{Introduction}
\label{sec:intro}
Machine learning has a profound impact on society, touching various aspects of our lives. It has gained significant influence in many fields, such as medical care, logistics, and knowledge dissemination. However, uncertainty in machine learning, which brings new risks and challenges to our society, disturbs its positive impact in more fields. Among many, we focus on privacy vulnerability. The public and related researchers' concerns about the privacy security of machine learning have increased as the ML technique plays an increasingly important role in our lives. Undoubtedly, safe and trustworthy data privacy preservation can build trust among the public and stakeholders, fostering the responsible and ethical use of machine learning technologies.

Despite the great advancement of machine learning, recent works \cite{chen2020ganmiarisk2,zhang2020ganmiarisk,yang2023rnnmiarisk} have shown a risk of privacy leakage in applications of machine learning models. Some studies \cite{shokri2017membership,salem2019nn,del2022leveraging} have successfully developed a proxy attacker to steal the membership privacy of a victim model by imitating the training process of the victim model. Besides, some other studies \cite{florian2016steal,kariyappa2021maze,truong2021dataextraction,sanyal2022towards} explored how to steal a well-trained model's information even without training data and knowing the model's architectural information. Despite the modern models' great performance, such potential risks make them difficult to apply in privacy-sensitive applications.

%In particular, compared to the reconstruction of privacy data (e.g., gradient-based data reconstruction attacks \cite{chen2021understanding, zhu2021rgap}), membership inference attacks (MIAs) have developed quite powerful. Its attack prerequisites, needing only prediction probabilities or even final predictions in most approaches, are usually easy to meet. In other words, MIAs pose a non-neglected threat in applications based on highly sensitive data. 
In particular, membership inference attack (MIA) is widespread due to its simple prerequisites that usually require only prediction probabilities or even final predictions. In contrast, other privacy attacks are harder to deploy in the real world because of their more restricted prerequisites (e.g., gradient-based data reconstruction attacks \cite{chen2021understanding,zhu2021rgap}).
The fundamentals of the membership inference attacks lie in the ability to distinguish the ML models' behaviors between on training and testing data that are also called `in' and `out' membership, or `member' and `non-member' data, respectively. For instance, the prediction confidence is significantly higher on `in' data than `out' data. %These behavioral differences widely exist in different fine-grained aspects: the different samples, the distorted sample and its original one, and the statistical distribution. %Inconsistencies in prediction results exist in different aspects: prediction results per se, prediction results under interfering conditions, gradients, etc.
Behavioral inconsistencies exist in different aspects, such as prediction confidence, robustness, etc.
Based on such discrepancies, an attacker can determine whether a sample is `in' membership or not.

Many existing defense mechanisms for ML models put lots of preconditions on the training procedure, such as oversaturated data or multiplying model costs. 
They have achieved privacy protection ability from different perspectives, e.g., unlearning partial data, aligning `in' and `out' predictions, and reducing the degree of fitting members. However, they still have some limitations.
For example, some studies \cite{nasr2018advreg,shejwalkar2021dmp} require a lot of additional data, which directly leads to a loss of models' generalizability. Or, some others \cite{abadi2016dpsgd,wang2021pruning,yuan2022samia,tang2022selena} require significantly more computational cost during the inference or training phases. 
The other group of work \cite{jia2019memguard} is not effective on every MIAs \cite{choquette2021labelonlymia}.
%can only be effective for certain settings of MIAs. 

The recent non-model-internal privacy defense approaches \cite{jia2019memguard,tang2022selena,yang2023purifier} using external modules to wrap the model and regenerate or decorate predictions, and they show promising competitiveness. However, from a long-term view, the defense that directly impacts the model's prediction distribution is worth exploring.
On one hand, obfuscation-based defense solutions \cite{jia2019memguard,tang2022selena,yang2023purifier} can provide effective privacy protection capabilities against certain types of attacks while the direct-model-impact solutions are general for MIAs since they consider the model distribution rather than specific attacks.
On the other hand, the direct-model-impact solution could lead to a better model design standard or training paradigm.
Besides, they could also be combined with any model-external privacy defense approaches to achieve better privacy-preserving ability.

Hence, in this paper, we propose a model-level plug-in solution against membership privacy attack, called \emph{Saturn Rings Classifier Module} (SRCM). The module focuses on the model's representation space itself so that it can directly affect the final prediction of both `in' and `out' data.
The main idea is to make `in' and `out' data indistinguishable by directly transforming the magnitude of hidden feature vectors of the two groups of data to specific ranges. By limiting the classification model's representation space, SRCM can significantly improve the model's privacy protection ability while maintaining generalizability. In summary, we make the following contributions:
\begin{itemize}
    \item To the best of our knowledge, our work is the first to study how the representation magnitude affects membership privacy leakage.%and direction of hidden features affect membership privacy leakage.
    \item Based on the observation, we propose \emph{Saturn Rings Classifier Module} that mitigates the privacy vulnerability with no loss of generalizability.
    \item We extensively evaluate SRCM and show that it not only can improve the privacy-preserving ability but also can be combined with non-model-internal level approaches (common existing approaches) to boost privacy protection.
\end{itemize}

\begin{figure*}[t]
     \centering
     \begin{subfigure}[]{0.24\linewidth}
        \centering
        \includegraphics[width=\linewidth]{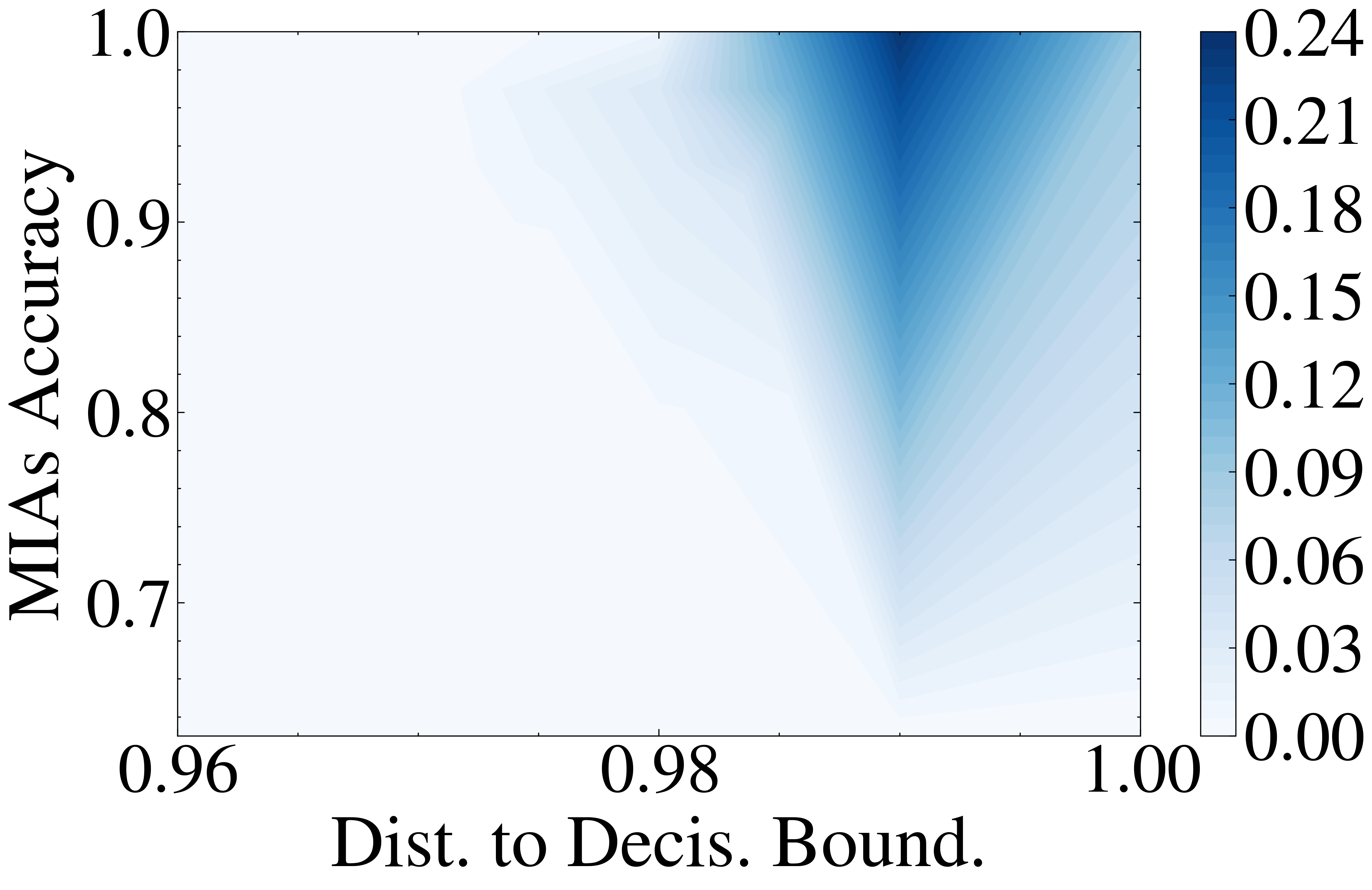}
        \caption{Member (train)}
        \label{fig:d2db_mia_train}
     \end{subfigure}
     \begin{subfigure}[]{0.24\linewidth}
        \centering
        \includegraphics[width=\linewidth]{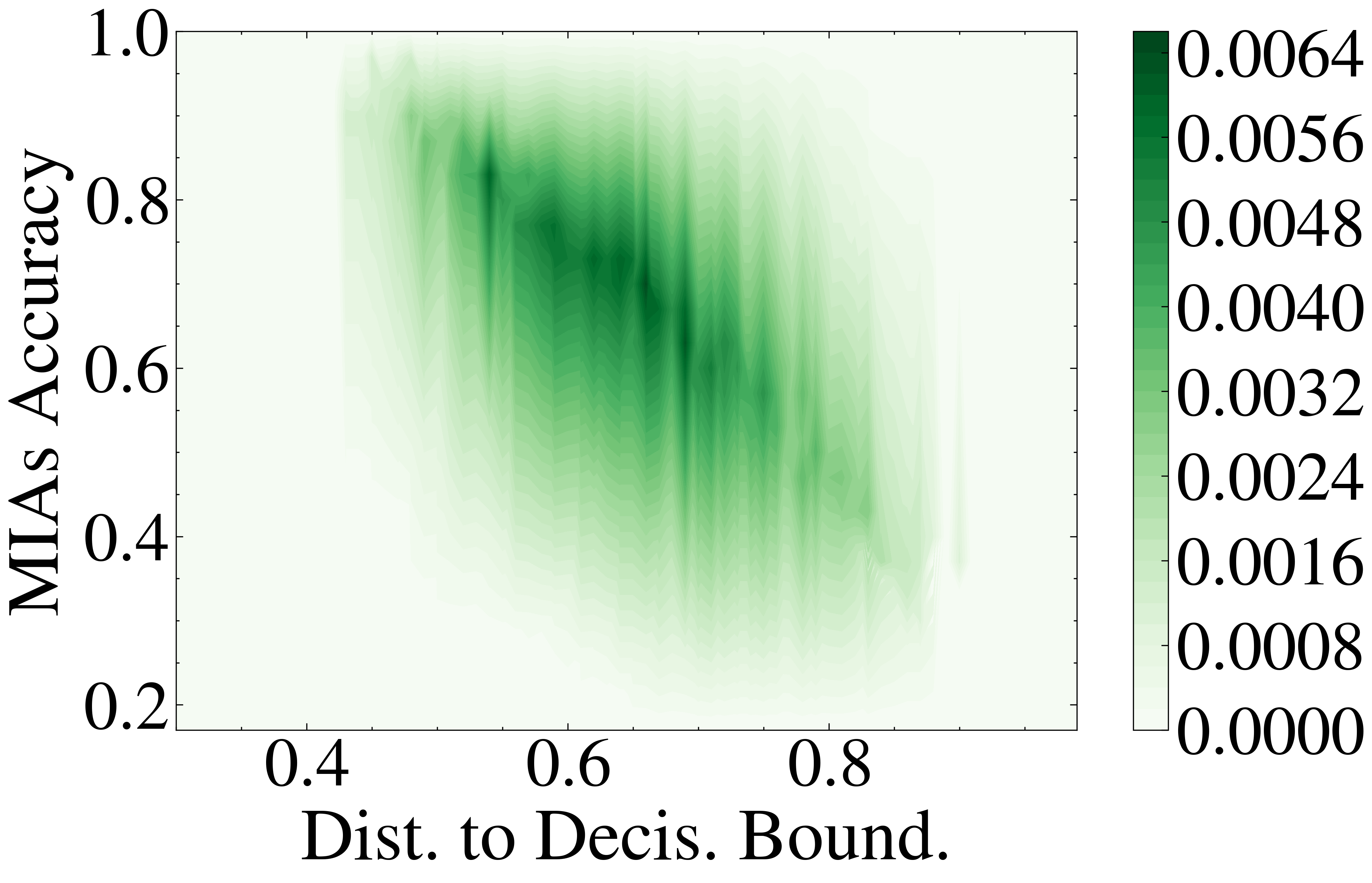}
        \caption{Non-Member (test)}
        \label{fig:d2db_mia_test}
     \end{subfigure}
     \begin{subfigure}[]{0.24\linewidth}
        \centering
        \includegraphics[width=\linewidth]{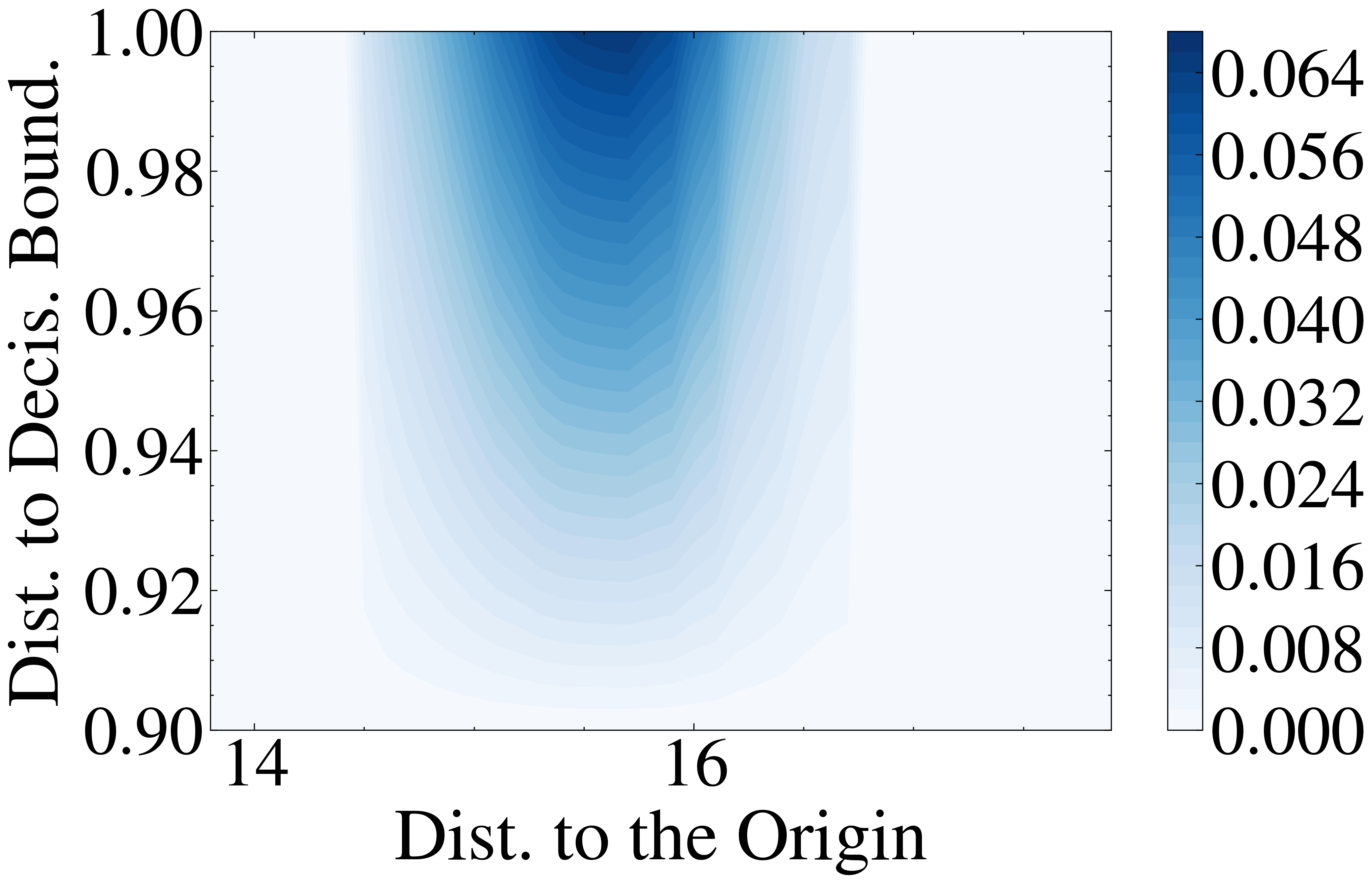}
        \caption{Member (train)}
        \label{fig:d2o_d2db_train}
     \end{subfigure}
     \begin{subfigure}[]{0.24\linewidth}
        \centering
        \includegraphics[width=\linewidth]{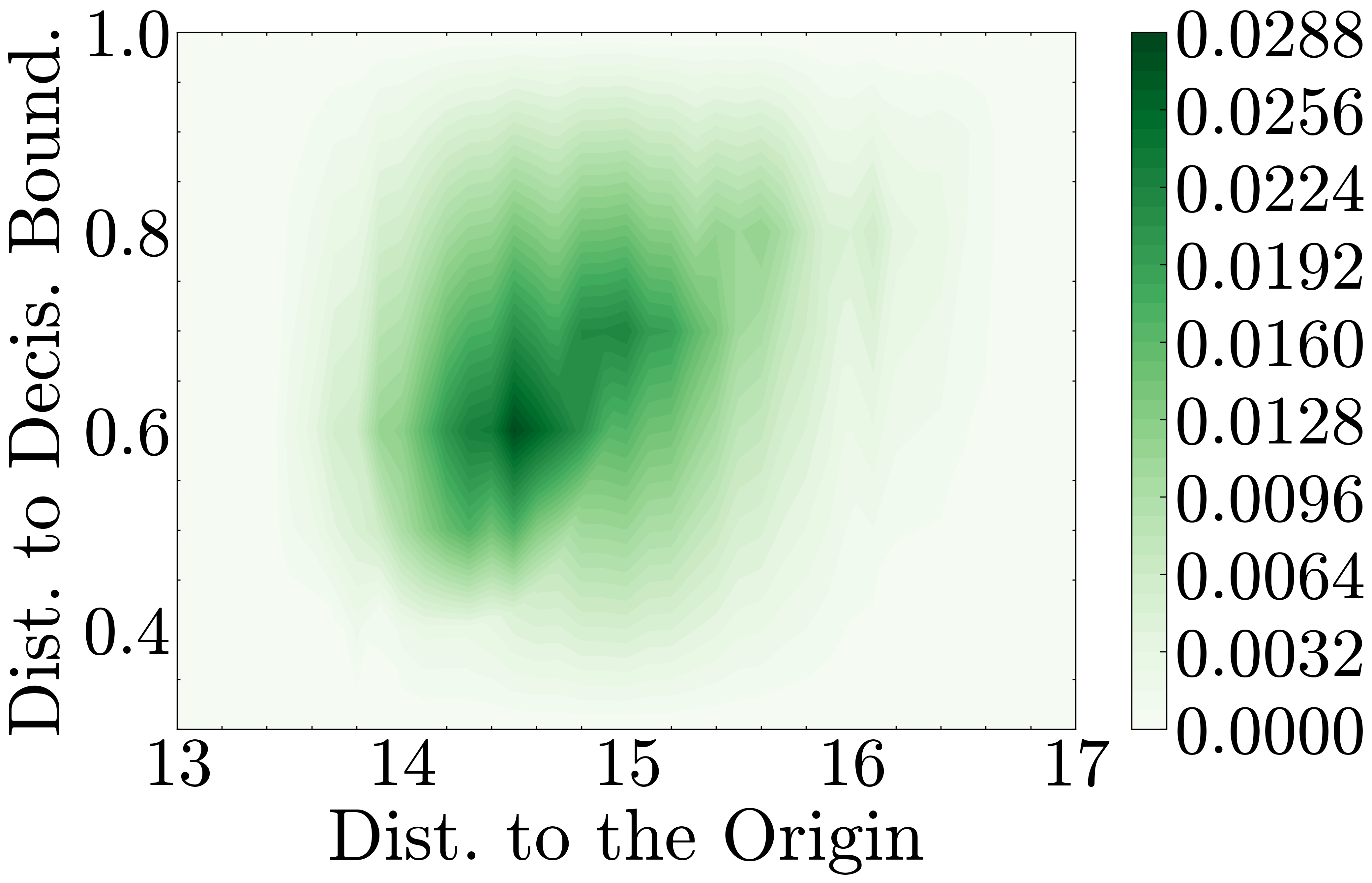}
        \caption{Non-Member (test)}
        \label{fig:d2o_d2db_test}
     \end{subfigure}
\caption{Relationship between the distance to the origin, the distance to the decision boundary, and MIAs accuracy. For a sample's distance to the decision boundary, we use the difference between 1st and 2nd maximum prediction probabilities as the metric. The four charts are obtained by averaging the results of dozens of independent experiments. The charts in \textcolor{blue}{blue} ((a) \& (c)) are produced on the training set, and the other charts in \textcolor{green}{green} ((b) \& (d)) are produced on the testing set. (ResNet18, CIFAR-100)}
\label{fig:d2o_d2db_mia}
\end{figure*}

\section{Related Work}
\label{sec:related_work}
\subsection{Membership Inference Attacks}
\cite{shokri2017membership} proposed the shadow model technique to simulate the prediction distribution of the target model. 
\cite{salem2019nn} proposed three kinds of attack approaches with lower-cost or less-preconditions. 
\cite{yuan2022samia} tried to add adversarial samples to enhance the MIAs.
Besides NN-based technique mentioned above that relies upon machine learning techniques to develop a proxy as the attacker, 
\cite{choquette2021labelonlymia} designed several label-only MIAs to reduce the attack requirements. 
\cite{song2021systematic} proposed modified cross-entropy (Mentr.) as a more sensitive attack metric.
\cite{del2022leveraging} introduced adversarial attacks into MIAs via the difference of the model predictions to a sample and its adversarial version.

\subsection{Privacy Defense with Direct Impact on Models}
Adversarial regularization \cite{nasr2018advreg} is one of the earliest studies that tried defense for MIAs via simulating the offensive behavior. 
\cite{salem2018mlleaks} proposed dropout and model stacking approaches to defend from MIAs.
Distillation for membership privacy (DMP) defense method \cite{shejwalkar2021dmp} selects `reference' data and generates proper labels to train a protected model.
\cite{hu2022defendinggan} tried to protect privacy by producing synthetic data through GAN.
\cite{yuan2022samia} explored pruning and found that it does not help the membership privacy, while \cite{wang2021pruning} concluded a contradiction.
\cite{chen2022relaxloss} were aware of the fitting priority problem and proposed RelaxLoss to help the model keep a trade-off between privacy and generalizability.
\cite{fang2024crl} proposed an approach to discriminate features among classes in the representation space while relaxing the model.
\cite{zhou2022frepo} tried to select the most valuable training data to help the model avoid memorizing the entire training set.
\cite{tarun2023dru} made the model unlearn the data in forgetting needs via alignment between models trained with and without sensitive data.
\cite{chundawat2023can} tried to make the student network forget specific sub-datasets through two differently trained teacher networks.

Prior studies have achieved great progress in privacy preservation via novel training diagrams or external decorators. However, discussion on the impact of architectures of neural networks remains insufficient. Hence, we study factors of modules in neural network architectures that impact privacy leakages in this paper.

\section{Problem Formulation}
In our study, we take into account classification tasks. In a scenario of membership inference attacks, an attacker basically tries to determine if a sample is a member of the training dataset by querying the target model (\textit{a.k.a.}, victim model). The membership inference attacks to a target model, $F_\theta: \mathbb{R}^{D_{in}} \rightarrow \mathbb{R}^{D_c}$, where $D_{in}$ and $D_c$ respectively denote the number of input features and task classes, can be formulated as:
\begin{equation}
    \mathcal{A}: \boldsymbol{x}, F_\theta \rightarrow \{0,1\},
\end{equation}
where $\mathcal{A}$ denotes a binary classification where the attacker outputs $1$ when the input $\boldsymbol{x}$ is predicted as a member of the training set used for the target model $F_\theta$, and $0$ otherwise.
The MIAs function $\mathcal{A}$ varies a lot according to the attack schemes. For NN-based MIAs, $\mathcal{A}$ is an ML model that uses the predictions of the target model as inputs. In contrast, when the attack scheme is based on some other metrics (such as threshold-based MIAs), the MIAs function is a manual set function
that computes the corresponding metrics and compares them with the threshold selected by statistical results via some techniques, such as using shadow models. 

\section{Methodology}
\label{sec:method}

\subsection{From Representation Space to Privacy}
To begin with, we ought to ask what makes the MIAs successful to gain the training membership information of the target model (\emph{a.k.a} victim model.) One consensus is that models behave differently on the members and non-members. However, the causes of the inconsistent behavior can vary. One of the causes can be attributed to the confidence gap. \cite{yuan2022samia} found out that class-level prediction confidence gaps often exist between member and non-member data in all classes. In a machine learning model, the degree of prediction confidence directly depends on how distant the sample's represented position of the logits in the model's representation space is from the decision boundary.

Then, does the disparity in the prediction results only come from the final classification layer? Clearly, the answer is no since if the classification layer receives similar inputs produced by the prior layers, then the prediction results should be similar. Due to logits being directly correlated to probabilities, their inevitable consistency makes it difficult to separate irrelevant and relevant factors for our goal. Another reason why the bottleneck layer matters is that it is effective in showing the overall distribution, and most network architectures have common bottleneck layers, naturally improving the universality of the empirical conclusion. Therefore, we move our attention to the deep features extracted from a bottleneck layer - usually the 2$^{nd}$ or 3$^{rd}$ last layer. A feature vector can be decomposed into magnitude (the distance to the origin) and direction (\textit{a.k.a.} angle). The angle, which directly determines if a sample is located within a class' decision area, has been widely explored as an important factor in classification models \cite{liu2016largemargin,wang2018cosface}. Unlike them, we question if the representation magnitude has such a significant relationship between decision boundaries and privacy. 

To further explore the relationship between distance to the decision boundary and MIAs accuracy, we visualize the sample-level distribution of the training and test sets. Fig.~\ref{fig:d2o_d2db_mia} shows the sample-level prediction results of MIAs accuracy vs. distance to the decision boundary, and distance to the origin vs. decision boundary. Trained with vanilla cross-entropy loss, the model's prediction and attack distributions differ on members and non-members.
On members, the model and the attacker are highly confident in all data. However, on non-members, the MIAs accuracy decreases when the sample becomes farther from the decision boundary, which indicates that the samples become closer to the training data. In Fig.~\ref{fig:d2o_d2db_train}, we find that the distance to the origin is uncorrelated to the distance to the decision boundary in members, while they are positively correlated (although not linearly) in non-member data as shown in Fig.~\ref{fig:d2o_d2db_test}.

We conjecture it is mainly due to the model using \emph{magnitude} to take \emph{shortcuts} in order to easily classify the members during training. 
Therefore, it is necessary to restrict the way of representation to guide the model to better representation in training and evaluation.
To achieve it, we propose \emph{Saturn Rings Classifier Module (SRCM)} in the next section.
%~\ref{sec:srcm}.

\subsection{Saturn Rings Classifier Module}
\label{sec:srcm}

Traditional activation functions (e.g., ReLU family \cite{agarap2019relu, djorkarn2016elu}) and normalization layers \cite{ioffe2015bn, ba2016layernormalization, ulyanov2017instancenormalization, wu2018groupnormalization} are mainly for solving overfitting and underfitting problems. However, they cannot avoid the occurrence of privacy leakage since they do not consider prediction alignment on the train and test sets. Unlike them, we propose the Saturn Rings Classifier Module (SRCM) to help the model's behavioral alignment in this section.

\begin{figure}[t]
  \centering
  \includegraphics[width=1.0\linewidth]{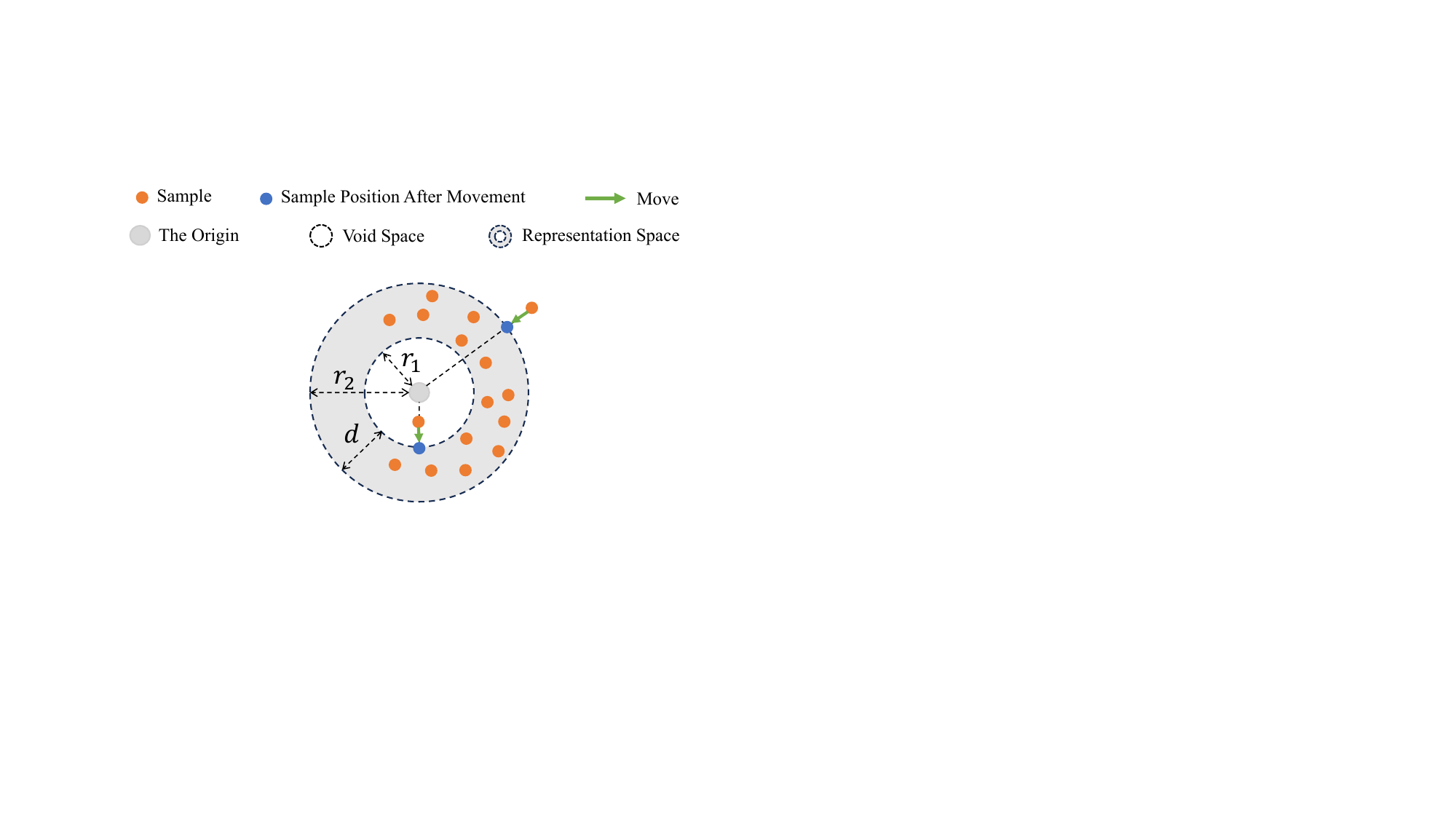}
  \caption{Illustration of Saturn rings activation function in 2D vector space. Our proposed method pushes samples into the representation space.}
  \label{fig:sraf}
\end{figure}

\subsubsection{Saturn Rings Activation Function (SR)}
To solve the under-confidence problem of non-member data, we designed the \emph{Saturn Rings Activation Function (SR)}.
As depicted in Fig.~\ref{fig:sraf}, the SR is composed of two n-sphere (2D in the figure) boundaries with radiuses $r_1$ and $r_2$ ($r_2>r_1$), respectively. The gap $d=r_2-r_1$. These boundaries delimit the representation space to the non-intersecting closed region they enclose, forming an annulus when $n=2$.
SR can be formulated as follows:
\begin{equation}
\label{eq:sr}
\mathtt{SR}(x, r_1, r_2)=
\begin{cases}
r_1 x / \|x\|, & \text{if }  \|x\| < r_1, \\
r_2 x / \|x\|, & \text{else if } \|x\| > r_2, \\
x, & \text{otherwise}\\
\end{cases}
\end{equation}
where $\|\cdot\|$ is the norm function. The boundary $r_1$ is intuitive that there should be a low boundary to guarantee that all samples, especially the non-members, stay far enough from the origin so that they are more likely to be predicted with higher confidence during evaluation.
However, the single boundary is practically ineffective when the model is in a high computational complexity (\emph{i.e.}, many more parameters or more complex connections.) Due to the huge computation requirement of these days' neural networks, models tend to push the training samples further away from the origin during training, leading to enlarging the gap between members and non-members.
Therefore, $r_2$ for an outer boundary is encouraged to construct a closed representation space.

To further explain the correlation relationship between the capacity of representation space $S_n$ ($n$ denotes the dimension of the vector space), we discuss its relationship between the inner boundary radius, $r_1$, and the outer boundary radius, $r_2$. 
We first establish their relationship within a 2D vector space and then generalize it to an n-dimensional vector space.
%For notation shorthand, we use $\Vec{g}$ to denote the classification model's last 2\nd layer's output of an input $x$. 
For a 2D vector space, $S_2$ can be formulated as follows:
\begin{equation}
\begin{split}
S_2 &= \pi(r_2^2-r_1^2) = \pi(r_2-r_1)(r_2+r_1) \\
    &= \pi d(r_1+r_2) \\ %= \pi d(d+2r_1) \\
    &\propto d(r_1+r_2) % = d(d+2r_1)
\end{split}
\end{equation}
Similarly, in 4D vector space,
\begin{equation}
\begin{split}
S_4 &= \frac{\pi^2}{2}d(r_1+r_2)(r_1^2+r_2^2) \\
    %&= \frac{\pi^2}{2}d(d+2r_1)(d^2+2dr_1+2r_1^2) \\
    &\propto d(r_1^{3} + r_1^{2}r_2^{1} + r_1^{1}r_2^{2} + r_2^{3})
\end{split}
\end{equation}
Accordingly, we find that the $S_n$ would be always proportional to the difference of powers between $r_1$ and $r_2$.
Using the `Difference of Powers Formula,' we can obtain their relationship in n-dimensional space: 
\begin{equation}
S_n \propto d \sum \limits_{i=0}^{n-1} r_1^{i}(r_1+d)^{n-1-i} %[(r_1+d)^{n-1} + (r_1+d)^{n-2}r_1 + \cdots + r_1^{n-1}]
\end{equation}
Due to the constraint that SR is supposed to limit the distance to the origin, $d$ should not be too large. Therefore, the capacity of the representation space is primarily determined by $r_1$. This indicates that, to maintain the representation space capacity, when we significantly reduce $r_1$, we only need to slightly increase $d$, and vice versa.

%\subsubsection{\makebox[.7\linewidth][s]{Magnitude Normalized Linear Layer (LinearNorm)}}
\begin{algorithm}[t]
\footnotesize
%\hspace{-5.cm}
%\linespread{0.7}
\SetAlgoLined
\PyComment{\makebox[.8\linewidth][s]{FC: vanilla fully connected layer class}} \\
\PyComment{fc: vanilla fully connected function} \\
\PyCode{\textcolor{blue}{class} LinearNorm(FC):}\\
\Indp   % start indent
    \PyCode{\textcolor{blue}{def} \_\_init\_\_(\textcolor{purple}{self}, all\_on, $\cdots$):}\\
    \Indp
        \PyCode{\textcolor{purple}{self}.w} \PyComment{weights inherited from FC} \\
        \PyCode{\textcolor{purple}{self}.b} \PyComment{bias inherited from FC} \\
        \PyCode{\textcolor{purple}{self}.all\_on} \PyComment{if norm all time} \\
    \Indm
    \PyCode{\textcolor{blue}{def} forward(\textcolor{purple}{self}, x):}\\
    \Indp
        \PyComment{x: outputs from the prior layer} \\
        %\PyComment{all\_on: if norm applied all time} \\
        \PyComment{if in training or SRCM mode} \\
        \PyCode{\textcolor{blue}{if} \textcolor{purple}{self}.training \textcolor{blue}{or} \textcolor{purple}{self}.all\_on:} \\
        \Indp
            \PyComment{normalize weights (l2-norm)} \\
            \PyCode{w = \textcolor{purple}{self}.w / \textcolor{purple}{self}.w.norm(p=2)} \\
        \Indm
        
        \PyCode{\textcolor{blue}{else}:}\\
        \Indp
            \PyCode{w = \textcolor{purple}{self}.w}\\
        \Indm
        \PyComment{perform forward pass} \\
        %\PyCode{o = fc(x, w, \textcolor{purple}{self}.b)} \\
        \PyCode{\textcolor{blue}{return} fc(x, w, \textcolor{purple}{self}.b)} \\
    \Indm 
\Indm % end indent, must end with this, else all the below text will be indented
%\caption{LinearNorm Pseudocode, PyTorch-like}
\caption{\makebox[.75\linewidth][s]{LinearNorm Pseudocode, PyTorch-like.}}
\label{algo:linear_norm}
\end{algorithm}

\subsubsection{Magnitude Normalized Linear Layer (LinearNorm)}
Although some studies \cite{liu2016largemargin,wang2018additive,wang2018cosface,deng2019arcface,sun2020circle,meng2021magface,kim2022adaface} were aware that the classification layer's weights' magnitude and direction have different impacts on the model's generalization ability in the training phase, research on the bottleneck layer's magnitude's impact on member and non-member data during training and evaluation is still unexplored. Here, we discuss how the bottleneck layer's outputs' magnitude interacts with the classification layer's weights.

There are two potential placements for SR: after or before the classification linear layer. If placed after the classification layer, SR can work individually, as it directly affects the magnitude of the logits. 
In contrast, if we simply apply SR prior to the standard classification layer, the classification layer may rescale the magnitude during training, resulting in SR being ineffective.
To avoid this situation, we propose \emph{Magnitude Normalized Linear Layer (LinearNorm)} (also see Algorithm \ref{algo:linear_norm}).
For the sake of notational simplicity, we use $\boldsymbol{g}$ to denote the classification model's last 2nd layer's output of an input $x$, $\boldsymbol{f}$ to denote the logits, and $W$ to denote the weights matrix, and $\boldsymbol{b}$ to denote a bias of the classification layer.
Therefore, the logits can be computed as follows:
\begin{equation}
\label{eq:f1}
\boldsymbol{f} = \boldsymbol{g}W + \boldsymbol{b}
               %= \|\boldsymbol{g}\| \|W\| \cos{\theta} + \boldsymbol{b}
\end{equation}
If $\boldsymbol{g}$ is a vector in $D_{h}$ dimensions, then $W$ is a $D_{h} \times D_{c}$ matrix for a $D_c$-class classification task. 
The $W$ can be decomposed to $[\boldsymbol{w_1}, \boldsymbol{w_2}, \cdots, \boldsymbol{w_{D_c}}]^\mathrm{T}$, where $\boldsymbol{w}$ is a vector with $D_h$ elements. Then, for the $i$-th class, we can obtain its corresponding logit as follows:
\begin{equation}
\label{eq:f2}
f_i = \boldsymbol{g}\boldsymbol{w}_i + b_i
    = \|\boldsymbol{g}\| \|\boldsymbol{w}_i\| \cos{\theta_i} + b_i
\end{equation}
where $\theta_i$ is the angle between vector $\boldsymbol{g}$ and $\boldsymbol{w}_i$, and $b_i$ is the $i$-th element of $\boldsymbol{b}$. Therefore, the logits depend on two factors: the magnitude and the direction. Although the magnitude of $\boldsymbol{g}$ is constrained by $r_1$ and $r_2$ in Eq.~\ref{eq:sr}, a learnable magnitude $\|\boldsymbol{w}_i\|$ could render this restriction ineffective, particularly for large models. Hence, a model with SR before the classification layer must normalize all $\boldsymbol{w}_i$s. Accordingly,
\begin{equation}
\label{eq:f3}
f_i = \boldsymbol{g}\boldsymbol{\hat{w}}_i + b_i
\end{equation}
where $\boldsymbol{\hat{w}}_i$ is a unit vector with the same direction of $\boldsymbol{w}_i$.
To simplify the computation, we use $\hat{W}$ = $W/\|W\|$ instead of $[\boldsymbol{\hat{w}_1}, \boldsymbol{\hat{w}_2}, \cdots, \boldsymbol{\hat{w}_{D_c}}]^\mathrm{T}$. Therefore, the LinearNorm layer can be represented as follows:
\begin{equation}
\label{eq:linearnorm}
\boldsymbol{f} = \boldsymbol{g}\hat{W} + \boldsymbol{b}
\end{equation}
An additional advantage using $W/\|W\|$ rather than $[\boldsymbol{\hat{w}_1}, \boldsymbol{\hat{w}_2}, \cdots, \boldsymbol{\hat{w}_{D_c}}]^\mathrm{T}$ is that it can achieve less accuracy loss since the module does not indicate all feature dimensions with the same importance (i.e., larger weights are regarded to carry more important features) while the fairness always requires trade-offs at the cost of accuracy \cite{wang2022instancefairness}.

\begin{figure}[t]
  \centering
  \includegraphics[width=1.\linewidth]{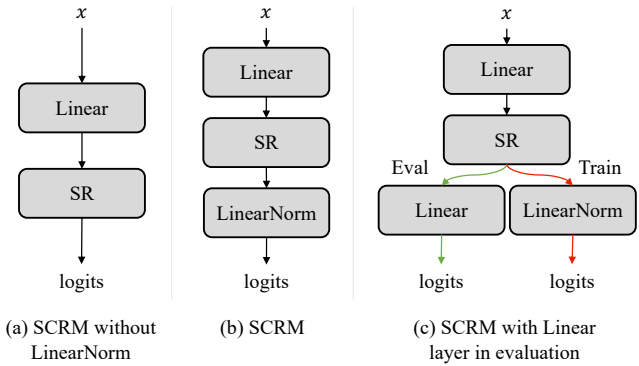}
  \caption{SCRM and variational designs for proof-of-concept purposes. $\rightarrow$ denotes common phase, \textcolor{green}{$\rightarrow$} denotes evaluation phase, and \textcolor{red}{$\rightarrow$} denotes training phase.}
  \label{fig:srcm}
\end{figure}
%\begin{figure}[t]
%     \centering
%     \begin{subfigure}[b]{0.23\linewidth}
%         \centering
%         \includegraphics[width=\linewidth]{draws/srcm_a.pdf}
%         \caption{CE (No defense)}
%         \label{fig:dist_density_vanilla}
%     \end{subfigure}
%     \hfill
%     \begin{subfigure}[b]{0.23\linewidth}
%         \centering
%         \includegraphics[width=\linewidth]{draws/srcm_b.pdf}
%         \caption{AdvReg ($\lambda_{adv}=1.0$)}
%         \label{fig:dist_density_advreg_1}
%     \end{subfigure}
%     \hfill
%     \begin{subfigure}[b]{0.5\linewidth}
%         \centering
%         \includegraphics[width=\linewidth]{draws/srcm_c.pdf}
%         \caption{AdvReg ($\lambda_{adv}=5.0$)}
%         \label{fig:dist_density_advreg_2}
%     \end{subfigure}
%\caption{}
%\label{fig:srcm}
%\end{figure}

%\subsubsection{Three Structural Designs}
\paragraph{Three Structural Designs}
We design three different structures as shown in Fig.~\ref{fig:srcm}. 
The comparison between Fig.~\ref{fig:srcm} (a) and (b) can justify where SR should be placed. Comparison between Fig.~\ref{fig:srcm} (b) and (c) can determine whether LinearNorm works in the evaluation stage. Note that there is an additional linear layer in Fig.~\ref{fig:srcm} (b) and (c) to mitigate the impact of existing various activation functions in various architectures.
For Fig.~\ref{fig:srcm} (b) and (c), a dual mode LinearNorm is implemented incorporating a switch shown in Algorithm~\ref{algo:linear_norm}.

\section{Experimental Setups}
\label{sec:exp_set}
\subsection{Datasets and Model Architectures}
Our approaches and others are evaluated on CIFAR-10, CIFAR-100 \cite{krizhevsky2009cifar100}, and Purchase100 \cite{purchase100}. CIFAR-10 and CIFAR-100 contain $60,000$ RGB images with a size of  $32\times32$. They have 10 and 100 classes, respectively. Purchase100 is a shopping dataset aiming to make appropriate discounts to attract shoppers to buy new products. Derived from a Kaggle challenge called `acquire valued shopper,' Purchase100 contains $197,342$ individuals' shopping records data. A simplified version in which each individual contains 600 binary features (every feature stands for a product) is applied in our study. All customers' records are labeled into $100$ classes.

In CIFAR-10, we normalize the original training and testing data as the common existing studies did. 
In CIFAR-100, data normalization, random flipping, and random cropping are applied to enhance the models' generalization ability and evaluate all approaches under these augmentations.
In Purchase100, we directly feed the original data into the models.
In all datasets, we first merge training and testing sets. Then, the whole dataset is split into two halves of equal quantity as target and shadow sets. To further make the results fair, all other settings, such as the optimizer and the way of splitting the dataset, remain the same throughout the experiments in each dataset. All experimental results are repeated five times or more (unless stated otherwise.)
The default random seed is set to 0 for reproducibility.

As for target models, we evaluate our approaches using VGG11 \cite{simonyan2015vgg}, ResNet18 \cite{he2016resnet}, and MobileNetV3 (large version) \cite{howard2019mobilenetv3} on CIFAR. For Purchase100, we apply a small multilayer perceptron (MLP) composed of four linear layers with hidden size $[1024,512,256]$ and Tanh activation function.

\subsection{Membership Inference Attacks}
In all MIAs approaches, the shadow model technique \cite{shokri2017membership} is applied. 
Some studies \cite{he2017ensemblenotstrong,athalye2018obfuscated} state that a perfect performance against attacks under ordinary conditions (\emph{a.k.a.} non-adaptive attacks) is not sufficient to claim that the defense approach is effective.
Hence, in this paper, adaptive attacks, which means the target model's training configurations and defense mechanisms are all known by the attacker, are also applied to evaluate various defense mechanisms' performance more accurately and conservatively. Besides the settings above, a neural network of four linear layers with hidden layer sizes [128, 64, 64] is applied when using \texttt{NN-based} MIAs. ReLU activation function and dropout technique are also applied to the attack model. Besides \texttt{NN-based} MIAs, we apply metric-based approaches, including Entropy-based method (\texttt{Entropy}) \cite{salem2018mlleaks}, Modified Entropy-based method (\texttt{M-Entropy}) \cite{song2021systematic}, and Gradient-based method with $\ell_2$ regularization (\texttt{Grad-$x$ $\ell_2$}) \cite{rezaei2021difficulty}.

\subsection{Defense Mechanism for MIA}
We compare our approach with a recent approach \texttt{RelaxLoss} \cite{chen2022relaxloss}, and a well-known approach, Adversarial regularization (\texttt{AdvReg}) \cite{nasr2018advreg}. Additionally, some other approaches, including \texttt{Early-stopping}, \texttt{Label-smoothing} \cite{guo2017labelsmoothing,Rafael2019labelsmoothing}, and \texttt{Confidence-penalty} \cite{pereyra2017confidencepenalty} are also included in our evaluation for comparison.
When using \texttt{AdvReg} approach, the settings follow the original paper to produce the inference attack model. 

\subsection{Common Configurations}
For all target and shadow models, stochastic gradient descent (SGD) optimizer with $0.09$ momentum and $5\times 10^{-4}$ weight decay is applied to the three datasets. To study the impact of magnitude in deep features, the last global average pooling layer or the 2$^{nd}$ last fully connected layer is chosen.

\newcolumntype{g}{>{\columncolor{myGray}}c}
\begin{table*}[t]
\centering
\resizebox{1.0\linewidth}{!}{
\large
\begin{tabular}{llgggggg|gggggg}%|ccccccc}
\toprule
& &
\multicolumn{6}{c}{\large{\textbf{CIFAR-10}}} & 
\multicolumn{6}{c}{\large{\textbf{CIFAR-100}}} \\%&
%\multicolumn{7}{c}{\large{\textbf{Purchase100}}} \\   
\cmidrule(lr){3-8} \cmidrule(lr){9-14} % \cmidrule(lr){17-23}
 \rowcolor{white}
\large{\textbf{Model}} & \large{\textbf{Train Tech.}}
& \large{\textbf{Train}} & \large{\textbf{Test}} & \large{\textbf{NN}} & \large{\textbf{Entr.}} & \large{\textbf{M-Entr.}} & \large{\textbf{Grad-$x$}}
& \large{\textbf{Train}} & \large{\textbf{Test}} & \large{\textbf{NN}} & \large{\textbf{Entr.}} & \large{\textbf{M-Entr.}} & \large{\textbf{Grad-$x$}}
%& \large{\textbf{Train}} & \large{\textbf{Test}} & \large{\textbf{MIAs 1}} & \large{\textbf{MIAs 2}} & \large{\textbf{MIAs 3}} & \large{\textbf{MIAs 4}} & \large{\textbf{MIAs 5}} 
\\
\midrule
%%%%%%%%%%%%%%%%%%%%%%%%%%%%%%%%%%%%%%%%%%%%%%
%%%%%%%%%%%%% MobileNetV3 (Large) %%%%%%%%%%%%
%%%%%%%%%%%%%%%%%%%%%%%%%%%%%%%%%%%%%%%%%%%%%%
\midrule  \rowcolor{white} 
& CE (baseline)
&100.00 & 74.83 & 80.26 & 72.53 & 73.62 & 73.52 
& 93.98 & 54.48 & 75.57 & 64.94 & 72.40 & 72.21 
%& 00.00 & 00.00 & 00.00 & 00.00 & 00.00 & 00.00
\\ \cmidrule(lr){2-14} \rowcolor{white}
& AdvReg
& 82.13 & 62.51 & 62.76 & 54.19 & 60.67 & 60.37
& 94.70 & 50.70 & 77.88 & 66.58 & 75.60 & 75.29
%& 00.00 & 00.00 & 00.00 & 00.00 & 00.00 & 00.00
\\ \cmidrule(lr){2-14} \rowcolor{white}
& RelaxLoss
& 79.84	& 71.43	& 59.80	& 55.77 & 57.56 & 58.92
& 83.40 & 54.47 & 69.54 & 60.32 & 67.60 & 68.08
%& 00.00 & 00.00 & 00.00 & 00.00 & 00.00 & 00.00
\\ \cmidrule(lr){2-14}
- (Large) & SCRM (Ours)
&100.00 & 74.63 & 77.10 & 70.32 & 71.78 & 71.74
& 91.91 & 54.39 & 71.41 & 63.14 & 71.06 & 70.64
%& 00.00 & 00.00 & 00.00 & 00.00 & 00.00 & 00.00
\\ \cmidrule(lr){3-14}
\multirow{-7}{*}{MobileNetV3}& + RelaxLoss
& 84.44 & 73.58 & 60.33 & 56.40 & 58.08 & 58.69
& 83.97 & 54.71 & 67.61 & 58.45 & 66.49 & 67.74
%& 00.00 & 00.00 & 00.00 & 00.00 & 00.00 & 00.00
\\
%%%%%%%%%%%%%%%%%%%%%%%%%%%%%%%%%%%%%%%%%%%%%%
%%%%%%%%%%%%%%% ResNet18 %%%%%%%%%%%%%%%%%%%%%
%%%%%%%%%%%%%%%%%%%%%%%%%%%%%%%%%%%%%%%%%%%%%%
\midrule \rowcolor{white}
%\multirow{7}{*}{ResNet18} 
& CE (baseline)
&100.00 & 70.31 & 88.09 & 85.91 & 86.44 & 86.31
&100.00 & 58.06 & 86.88 & 82.96 & 84.04 & 84.20
%& 00.00 & 00.00 & 00.00 & 00.00 & 00.00 & 00.00
\\ \cmidrule(lr){2-14} \rowcolor{white}
& AdvReg
& 99.95 & 61.89 & 81.58 & 74.99 & 78.74 & 77.57
& 94.24 & 47.94 & 76.11 & 72.36 & 81.45 & 72.06
%& 00.00 & 00.00 & 00.00 & 00.00 & 00.00 & 00.00
\\ \cmidrule(lr){2-14} \rowcolor{white}
& RelaxLoss
& 91.56 & 69.25 & 77.32 & 71.51 & 72.25 & 73.50
& 91.78 & 57.97 & 78.22 & 73.04 & 75.38 & 76.13
%& 00.00 & 00.00 & 00.00 & 00.00 & 00.00 & 00.00
\\ \cmidrule(lr){2-14}
& SCRM (Ours)
& 100.00 & 71.43 & 80.35 & 82.17 & 82.78 & 82.67
& 92.86 & 57.91 & 50.00 & 71.65 & 73.38 & 77.05
%& 00.00 & 00.00 & 00.00 & 00.00 & 00.00 & 00.00
\\ \cmidrule(lr){3-14}
\multirow{-7}{*}{ResNet18} & + RelaxLoss
& 92.00 & 70.52 & 74.28 & 69.77 & 70.53 & 71.89
& 92.22 & 58.58 & 58.84 & 61.48 & 70.38 & 76.89
%& 00.00 & 00.00 & 00.00 & 00.00 & 00.00 & 00.00
\\
%%%%%%%%%%%%%%%%%%%%%%%%%%%%%%%%%%%%%%%%%%%%%%
%%%%%%%%%%%%%%%%%%% VGG11 %%%%%%%%%%%%%%%%%%%%
%%%%%%%%%%%%%%%%%%%%%%%%%%%%%%%%%%%%%%%%%%%%%%
\midrule \rowcolor{white}
& CE (baseline)
& 100.00 & 76.46 & 76.31 & 74.15 & 74.90 & 75.33
& 99.97 & 53.75 & 85.18 & 81.67 & 83.08 & 83.46
%& 00.00 & 00.00 & 00.00 & 00.00 & 00.00 & 00.00
\\ \cmidrule(lr){2-14} \rowcolor{white}
& AdvReg
& 99.28 & 69.52 & 72.43 & 64.96 & 69.22 & 69.84
& 99.91 & 50.08 & 89.20 & 87.11 & 89.19 & 77.00
%& 00.00 & 00.00 & 00.00 & 00.00 & 00.00 & 00.00
\\ \cmidrule(lr){2-14}  \rowcolor{white}
& RelaxLoss
& 94.44 & 75.88 & 69.64 & 66.87 & 66.87 & 68.12
& 91.77 & 53.42 & 76.94 & 72.27 & 76.08 & 76.03
%& 00.00 & 00.00 & 00.00 & 00.00 & 00.00 & 00.00
\\ \cmidrule(lr){2-14}
& SCRM (Ours)
& 100.00 & 75.69 & 57.09 & 72.96 & 73.37 & 74.35
& 98.81 & 53.67 & 82.90 & 76.95 & 80.37 & 81.52
%& 00.00 & 00.00 & 00.00 & 00.00 & 00.00 & 00.00
\\ \cmidrule(lr){3-14} 
\multirow{-7}{*}{VGG11} & + RelaxLoss
& 95.70 & 75.07 & 61.46 & 66.14 & 66.62 & 67.66
& 89.55 & 52.79 & 75.16 & 69.81 & 73.07 & 75.29
%& 00.00 & 00.00 & 00.00 & 00.00 & 00.00 & 00.00
\\
%\midrule
%\midrule
%%%%%%%%%%%%%% details
\bottomrule
\end{tabular}
} % \resizebox
\caption{
    \textbf{Evaluations on CIFAR-10, and -100} -- `Train' and `Test' stand for training and testing accuracy, respectively. All MIAs are reported in AUC scores.
} % \caption
\label{tab:exp_ac_res_3}%
\end{table*}
\begin{figure}[t]
  \centering
  \includegraphics[width=.8\linewidth]{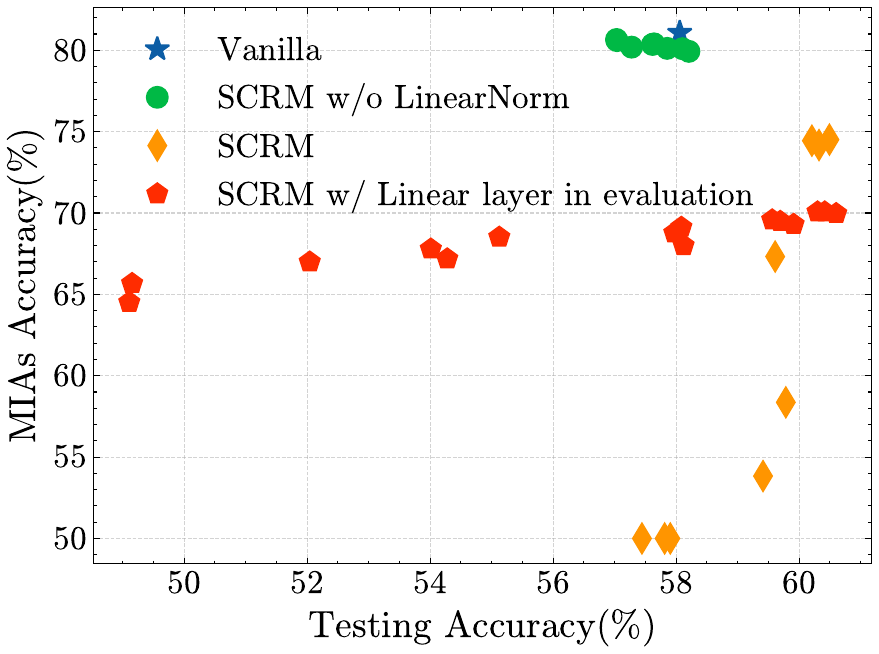}
  %\vspace*{-3mm}
  \caption{Comparison of our proposed SRCM and two other variational designing options for proof of concept - one without LinearNorm, and the other with a Linear layer in the evaluation phase. `Vanilla' denotes the original model (baseline). Rightward (higher testing accuracy) and lower (lowe MIA accuracy) is better. (ResNet18, CIFAR-100)}
  \label{fig:cmp_srcm_acc_mia}
\end{figure}
\begin{figure}[]
     \centering
     \begin{subfigure}[]{0.32\linewidth}
        \centering
        \includegraphics[width=\linewidth]{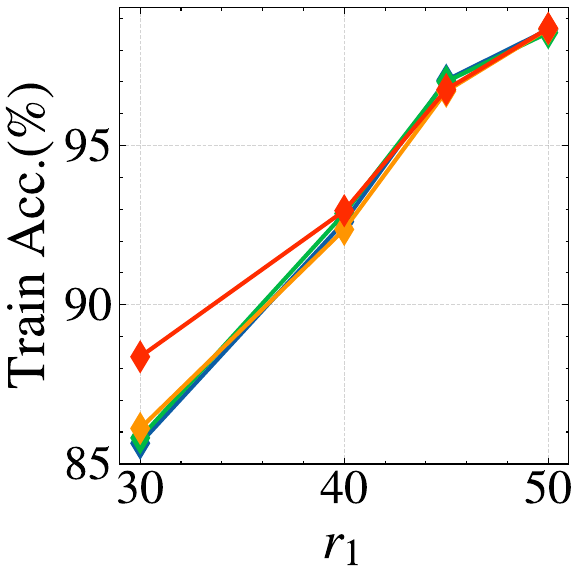}
        %\vspace*{-5mm}
        \caption{Train}
        \label{fig:srcm_train}
     \end{subfigure}
     \begin{subfigure}[]{0.32\linewidth}
        \centering
        \includegraphics[width=\linewidth]{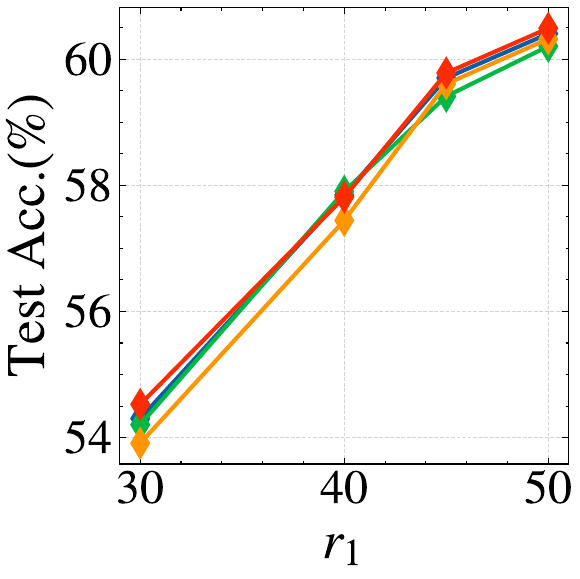}
        %\vspace*{-5mm}
        \caption{Test}
        \label{fig:srcm_test}
     \end{subfigure}
     \begin{subfigure}[]{0.32\linewidth}
        \centering
        \includegraphics[width=\linewidth]{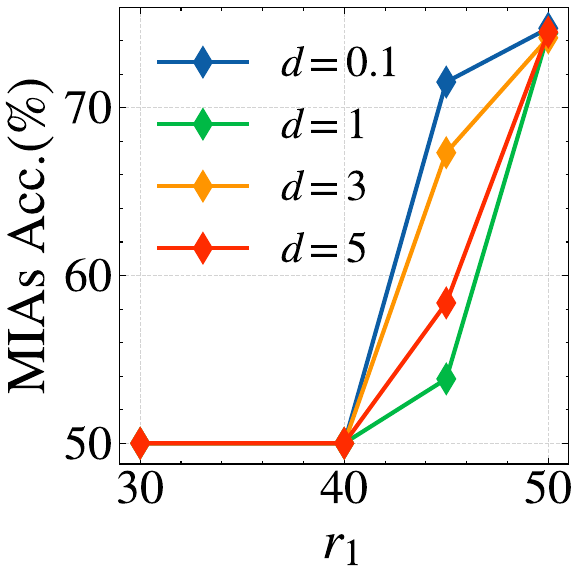}
        %\vspace*{-5mm}
        \caption{MIAs}
        \label{fig:srcm_mia}
     \end{subfigure}
     %\vspace*{-2.mm}
\caption{Training, testing, and MIAs accuracy changes with various hyper-parameters combinations using SRCM. (ResNet18, CIFAR-100)}
\label{fig:srcmb_conf}
\end{figure}

%%%%%%%%%%%%%%%%%%%%%%%%%%%%%%%%%%%%%%%%%%%%%%%%%%%%%%%%%%%%%%
\section{Empirical Study}
\label{sec:exp}
\begin{figure*}[th]
    \centering
    \includegraphics[width=1.0\linewidth]{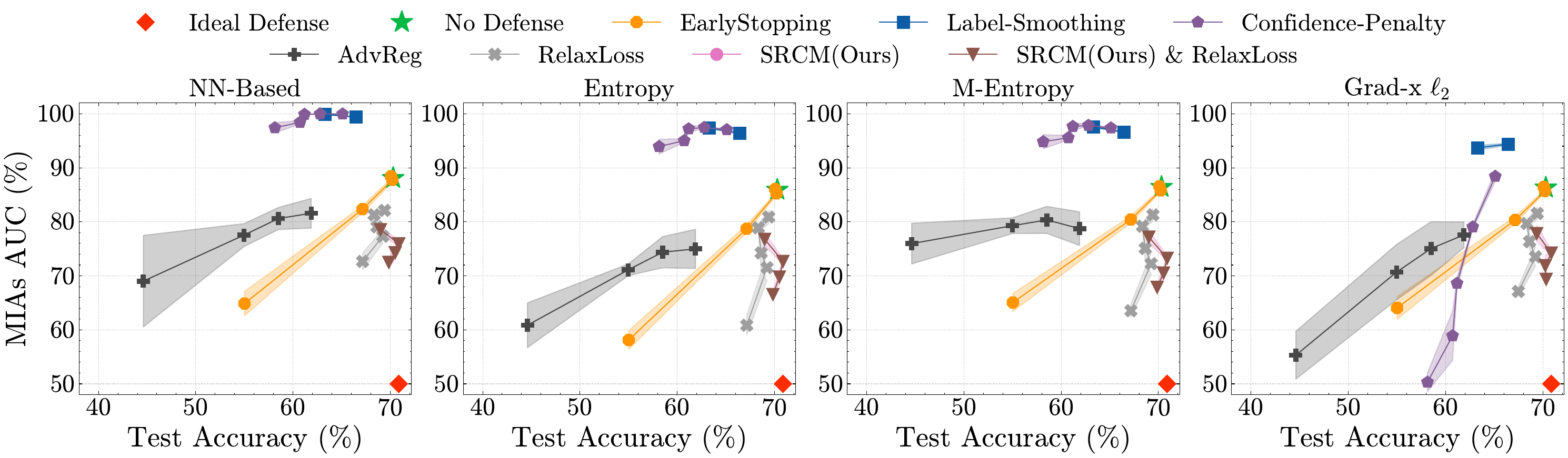}
    \caption{Performance of defenses against different adaptive MIAs (ResNet18, CIFAR-10). Being lower and more rightward is better.}
    \label{fig:acc_mia_c10}
\end{figure*}
\begin{figure*}[th]
    \centering
    \includegraphics[width=1.0\linewidth]{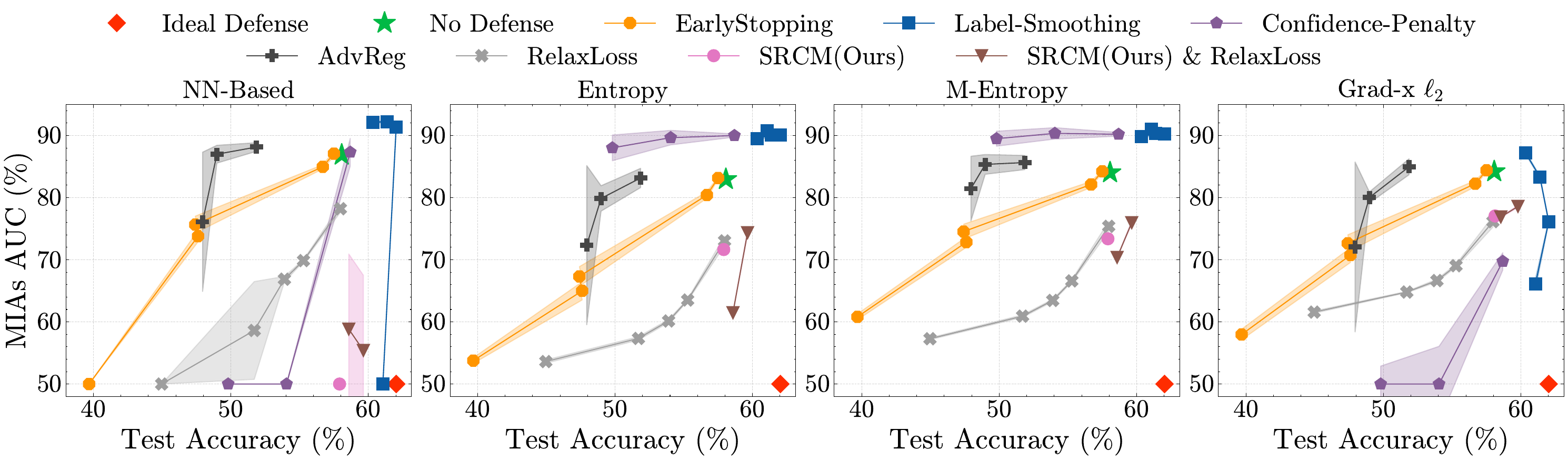}
    \caption{Performance of defenses against different adaptive MIAs (ResNet18, CIFAR-100). Being lower and more rightward is better.}
    \label{fig:acc_mia_c100}
\end{figure*}
\begin{figure*}[t]
    \centering
    \includegraphics[width=1.0\linewidth]{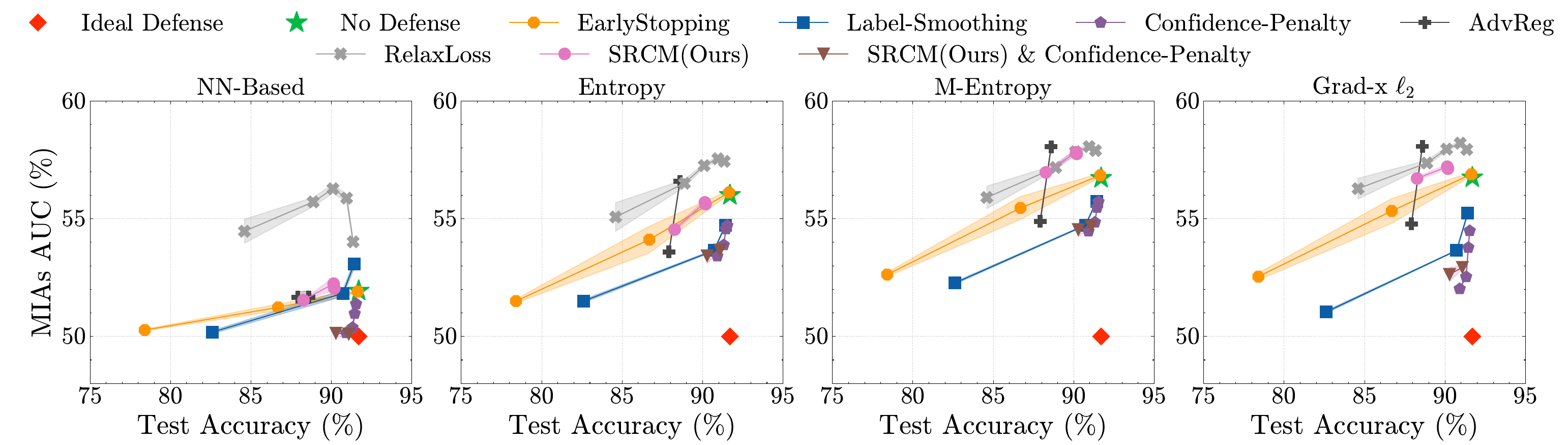}
    \caption{Performance of defenses against different adaptive MIAs (MLP, Purchase100). Being lower and more rightward is better.}
    \label{fig:acc_mia_p100}
\end{figure*}

\subsection{Comparison between Three Designs}
%To reduce redundant comparisons, it would be better to verify our three designs' effectiveness before comparing them with other existing methods. 
%First, we verify the effectiveness of the three design choices, SRCM-A, -B, and -C, as shown in Fig.~\ref{fig:srcm}.

We verify our design of SRCM with other variational designing options for proof of concept - one without LinearNorm and the other with a Linear layer in the evaluation phase.
Firstly, we design multiple sets of hyper-parameters and show the relationship between testing accuracy and MIAs accuracy. Shown in Fig.~\ref{fig:cmp_srcm_acc_mia}, we show the impacts of three designs on testing accuracy and MIAs accuracy. It shows that SRCM can preserve privacy best, meaning that the common computing process in the training and testing phase helps the protection of privacy. 
The results in Fig.~\ref{fig:cmp_srcm_acc_mia} are based on the grid search of hyper-parameters combinations. The three structures show the effectiveness of SR and LinearNorm. 
Omitting LinearNorm layer from SRCM may not be sufficiently effective on ResNet18 because placing it on the end of the model will fully lose its adaptive ability to adjust the magnitude of logits. Meanwhile, it also verifies the effectiveness and necessity of LinearNorm.

%After showing that SRCM-B is a better way for privacy preservation, the configurations in which SRCM-B can be effective need to be discussed. 
The SRCM has been validated as an effective way for privacy preservation. Then, we explore how it changes with various hyper-parameter combinations (radius $r_1$ and radius gap $d$).
As shown in Fig.~\ref{fig:srcmb_conf}, we explore the relationship between performance (privacy and accuracy) and hyper-parameters (inner radius $r_1$ and radius gap $d$). We find that a model's representation space capacity impacts its generalizability and memorizing ability. All three figures of Fig.~\ref{fig:srcmb_conf} agree that training, testing, and MIAs accuracy are all positively correlated with $r_1$. This is aligned with our expectation that the larger the capacity of the feature space is, the greater the degree of representation freedom of the model itself.

From Fig.~\ref{fig:srcm_train} and Fig.~\ref{fig:srcm_test}, we note that, unlike network pruning, reducing network capacity by SRCM has a more consistent impact across training and testing accuracy. (In network pruning, either in unstructured magnitude or structured pruning, there is a discrepancy in training and testing accuracy.) This indicates that changing the representation space's capacity and shape significantly affects the model's \emph{memorizing} ability. Since the computation capacity of the other hidden layers is not directly reduced (while network pruning directly reduces the compute capacity), the model can still maintain generalizability. Combined with Fig.~\ref{fig:srcm_mia}, we can find that a proper range of magnitude can maintain a better accuracy and privacy trade-off. In particular, we empirically found that a `sweet spot' (that is shown in the following subsection) exists in a model's representation space, allowing the model to gain more privacy with no or little accuracy loss. In other words, this means that a larger range of representation magnitude may achieve limited accuracy improvement while resulting in a significant loss of privacy-preserving ability. It suggests that further reducing the range will result in a greater loss of accuracy with insignificant privacy improvement. It also verifies that a model requires sufficient computational capacity to support its generalizability.

\newcolumntype{g}{>{\columncolor{myGray}}c}
\begin{table*}[th]
\centering
\resizebox{0.75\linewidth}{!}{
\large
\begin{tabular}{ll|gg|gggg}%|ccccccc}
\toprule \rowcolor{white}
\large{\textbf{Model}} & \large{\textbf{Train Tech.}}
& \large{\textbf{Train Acc.}} & \large{\textbf{Test Acc.}} & \large{\textbf{NN}} & \large{\textbf{Entr.}} & \large{\textbf{M-Entr.}} & \large{\textbf{Grad-$x$}}
\\
\midrule
%%%%%%%%%%%%%%%%%%%%%%%%%%%%%%%%%%%%%%%%%%%%%%
%%%%%%%%%%%%%%%%% Purchase100 %%%%%%%%%%%%%%%%
%%%%%%%%%%%%%%%%%%%%%%%%%%%%%%%%%%%%%%%%%%%%%%
\midrule \rowcolor{white}
& CE (baseline)
& 99.98 & 91.70 & 51.94 & 56.00 & 56.73 & 56.75 
\\ \cmidrule(lr){2-8} \rowcolor{white}
& Early-Stopping
& 95.50 & 86.68 & 51.23 & 54.11 & 55.45 & 55.33
\\ \cmidrule(lr){2-8} \rowcolor{white}
& Label-Smoothing
& 97.94 & 90.71 & 51.83 & 53.65 & 54.72 & 53.65
\\ \cmidrule(lr){2-8} \rowcolor{white}
& Confidence-Penalty
& 97.04 & 90.91 & 50.14 & 53.41 & 54.47 & 52.02
\\ \cmidrule(lr){2-8} \rowcolor{white}
& AdvReg
& 99.96 & 88.59 & 51.66 & 56.59 & 58.05 & 58.07
\\ \cmidrule(lr){2-8} \rowcolor{white}
& RelaxLoss
& 97.24 & 88.86 & 55.71 & 56.50 & 57.17 & 57.36
\\ \cmidrule(lr){2-8}
& SCRM(Ours)
& 98.80 & 90.18 & 52.04 & 55.62 & 57.75 & 57.12
\\ \cmidrule(lr){3-8}
\multirow{-11}{*}{MLP} & +Confidence-Penalty
& 96.97 & 90.28 & 50.14 & 53.43 & 54.54 & 52.63
%\\ \cmidrule(lr){3-8}
%& +Label-Smoothing
%& 00.00 & 00.00 & 00.00 & 00.00 & 00.00 & 00.00 
\\
%%%%%%%%%%%%%% details
\bottomrule
\end{tabular}
} % \resizebox
\caption{
    \textbf{Evaluations on Purchase100} -- All MIA approaches are reported in AUC scores.
} % \caption
\label{tab:exp_ac_res_p100}%
\end{table*}

\subsection{More Results and Discussion}
In CIFAR-10, as shown in Table.~\ref{tab:exp_ac_res_3}, the combination of SRCM and RelaxLoss achieves significant improvement over others. Through experiments on MobileNetV3, SRCM has achieved significant improvements in privacy preservation with minimal accuracy loss. Also, combining with RelaxLoss provides further improvement. Experiments on ResNet18 also show a similar trend. However, trends slightly changed when we conducted the experiments on VGG11. The effect of SRCM becomes less effective than that on other larger neural networks. We attribute it to the insufficient computation capacity of the model, which is verified by the experiments with Purchase100 in Fig.~\ref{fig:acc_mia_p100}. Additional experiments on ResNet18 are shown in Fig.~\ref{fig:acc_mia_c10}. In the figure, Label-smoothing and Confidence-Penalty have a significantly negative impact on the model under NN-Base and two Entropy-based attacks. Interestingly, Confidence-Penalty has a strong defensive effect against Grad-$x$ attacks. Label-Smoothing does not have a positive effect in all four MIAs setups. EarlyStopping, AdvReg, and RelaxLoss perform better than the other approaches. In particular, RelaxLoss outperforms the other two approaches. %Although SRCM does not perform better than RelaxLoss, it can be applied with other approaches.
With the combination of RelaxLoss, our approach is even further enhanced since SRCM and RelaxLoss optimize privacy from different perspectives.

In CIFAR-100, similar trends are observed. Compared to that in CIFAR-10, the effectiveness of all methods diminishes as task difficulty increases. As shown in Fig.~\ref{fig:acc_mia_c100}, the impact of \texttt{AdvReg} becomes more limited, especially on NN-Base and M-Entropy. Both SCRM and the combination of SCRM and RelaxLoss achieve outstanding results. One thing to note is that SCRM becomes less effective if the method significantly degrades the model performance (the model must not lose its learnability).

In Purchase100, the trends significantly change. As shown in Table~\ref{tab:exp_ac_res_p100}, the MIAs AUC scores become significantly lower due to the smaller gap between testing and training sets. 
As a lightweight network with much fewer parameters than other networks in this study, it achieves significant improvement with Confidence-Penalty and Label-Smoothing over other approaches. This reflects the difference between shallow neural networks and DNN. Due to the weaker fitting ability of the model compared to DNN, RelaxLoss has little effect in this scenario. AdvReg's regularization mechanism is similar to the Confidence-Penality and starts to achieve the effect after losing some accuracy caused by splitting the training set.
Additional results are presented in Fig.~\ref{fig:acc_mia_p100}. Unlike the results in CIFAR, RelaxLoss exhibits a discontinuous trend, which makes it less effective in privacy preservation. Aligned with our hypothesis, SRCM does not achieve significant performance enhancement on such a small model due to the low computation capacity.

\begin{table}[t]
\small
  \centering
  \resizebox{1.0\linewidth}{!}{
  \begin{tabular}{c|cccc}
    \toprule
      Classifier & ResNet18 & VGG11 & MobileNetV3-Large\\
    \midrule 
        Vanilla & 17.80($\pm$0.07) & 5.23($\pm$0.00) & 8.65($\pm$0.10)  \\
        SRCM    & 17.81($\pm$0.10) & 5.35($\pm$0.02) & 8.83($\pm$0.20)  \\
    \bottomrule
  \end{tabular}
  }
  \caption{The latency (ms) comparison among different models w/ or w/o SRCM. (Run with AMD Ryzen 7 7700X and NVIDIA GeForce RTX 3080)}
  \label{tab:cmp_inf_latency}
\end{table}

For efficiency evaluation, we measure a series of models with and without SRCM to show the computational cost of our approach. Seen in Table.~\ref{tab:cmp_inf_latency}, SRCM shows minor inference time increase. Compared with the model's original inference cost, the difference is insignificant.

In summary, SRCM has a significant privacy-preserving impact on the models with sufficient computing capacity and can incorporate many existing methods to achieve cooperative privacy enhancement. Its effectiveness in those models suggests that redundant computing capacity can potentially be converted into privacy-preserving capabilities to some extent.

\section{Conclusion}
\label{sec:conclusion}
 We explored the privacy-leaking factors and presented a lightweight yet effective neural network component, \texttt{SRCM}, which mitigates the privacy vulnerability of over-parameterized classification models by restricting models' representation capacity. The insight of this work is that privacy vulnerability can be mitigated by aligning the factors that are in disparities between members and non-members. Through experiments, we validated our hypothesis and the effectiveness and ease of use of our approach. Importantly, we found new possibilities for making use of the models' oversaturated computation capacity. %We expect more unknown characteristics of ML models to be further excavated and utilized.

\bibliography{refs}

\end{document}